\documentclass[10pt,journal,compsoc]{IEEEtran}

\ifCLASSOPTIONcompsoc
  \usepackage[nocompress]{cite}
\else
  \usepackage{cite}
\fi

\usepackage{amsmath}
\usepackage{amssymb}
\usepackage{graphicx}
\usepackage{booktabs}       
\usepackage{algorithm}
\usepackage{algorithmicx}
\usepackage{algpseudocode}
\usepackage{subfigure}
\usepackage{url}
\usepackage[switch]{lineno}

\ifCLASSINFOpdf
\else
\fi

\newcommand{\dom}[1]{\mathcal{D}_{#1}}

\begin{document}
%
\title{ Exploring Explicit Domain Supervision for Latent Space Disentanglement in Unpaired Image-to-Image Translation}
%
%
%
%

\author{Jianxin Lin, ~Zhibo Chen,~\IEEEmembership{Senior~Member,~IEEE}, ~Yingce Xia, ~Sen Liu, \\
         Tao Qin, and ~Jiebo Luo,~\IEEEmembership{Fellow,~IEEE}
\IEEEcompsocitemizethanks{\IEEEcompsocthanksitem Zhibo Chen (Corresponding Author), Jianxin Lin and Sen Liu are with University of Science and Technology of China, Hefei, Anhui, 230026, China, (e-mail: chenzhibo@ustc.edu.cn).\protect\\

\IEEEcompsocthanksitem Yingce Xia and Tao Qin are with Microsoft Research Asia, Beijing, 100080, China, (e-mail: Yingce.Xia@microsoft.com and taoqin@microsoft.com).\protect\\
\IEEEcompsocthanksitem Jiebo Luo is with the Department of Computer Science, University of Rochester, Rochester, NY 14627 USA, (e-mail: jiebo.luo@gmail.com).}
\thanks{This work was supported in part by NSFC under Grant 61571413, 61632001.}}

%
%

\markboth{}%
{Shell \MakeLowercase{\textit{et al.}}: Bare Demo of IEEEtran.cls for Computer Society Journals}
%



\IEEEtitleabstractindextext{%
\begin{abstract}
Image-to-image translation tasks have been widely investigated with Generative Adversarial Networks (GANs). However, existing approaches are mostly designed in an unsupervised manner while little attention has been paid to domain information within unpaired data. In this paper, we treat domain information as explicit supervision and design an unpaired image-to-image translation framework, Domain-supervised GAN (DosGAN), which takes the first step towards the exploration of explicit domain supervision. In contrast to representing domain characteristics using different generators or domain codes, we pre-train a classification network to explicitly classify the domain of an image. After pre-training, this network is used to extract the domain-specific features of each image. Such features, together with the domain-independent features extracted by another encoder (shared across different domains), are used to generate image in target domain. Extensive experiments on multiple facial attribute translation, multiple identity translation, multiple season translation and conditional edges-to-shoes/handbags demonstrate the effectiveness of our method. In addition, we can transfer the domain-specific feature extractor obtained on the Facescrub dataset with domain supervision information to unseen domains, such as faces in the CelebA dataset. We also succeed in achieving conditional translation with any two images in CelebA, while previous models like StarGAN cannot handle this task.
\end{abstract}

\begin{IEEEkeywords}
Image-to-image translation, explicit domain supervision, generative adversarial networks.
\end{IEEEkeywords}}

\maketitle

\IEEEdisplaynontitleabstractindextext

%
\IEEEpeerreviewmaketitle

\IEEEraisesectionheading{\section{Introduction}\label{sec:introduction}}
Image-to-image translation covers a wide variety of computer vision problems, including image stylization~\cite{gatys2016image}, segmentation~\cite{long2015fully} and image super-resolution~\cite{ledig2016photo}. It aims at learning a mapping that can convert an image from a source domain to a target domain, while preserving the main presentations of the input images. For example, in the aforementioned three tasks, an input image might be converted to a portrait similar to Van Gogh's styles, a segmentation map splitted into different regions, or a high-resolution image, while the content remains unchanged. Since it is usually challenging to collect a large amount of paired data for such tasks, unsupervised learning algorithms have been widely adopted in image-to-image translation models. Particularly, the generative adversarial networks (GAN)~\cite{goodfellow2014generative} and dual learning~\cite{he2016dual} are studied. One of the most popular model, CycleGAN \cite{zhu2017unpaired}, tackles cross-domain unpaired image-to-image translation by the aforementioned two techniques, where the GANs are used to ensure the generated images belonging to the target domain, and dual learning can help improve image qualities by minimizing reconstruction loss.

An implicit assumption of image-to-image translation is that an image contains two kinds of features\footnote{ Note that the two kinds of features are relative concepts, and domain-specific features in one task might be domain-independent features in another task, depending on what domains one focuses on in the task.} \cite{lin2018conditional,huang2018multimodal,lee2018diverse}: {\em domain-independent features}, which are preserved during the translation (i.e., the content when translating a natural image to Van Gogh' styles), and {\em domain-specific features}, which are changed during the translation (i.e., the styles when translating the image to Van Gogh' styles).  Image-to-image translation aims at transferring images from the source domain to the target domain by preserving domain-independent features while replacing domain-specific features. Therefore, the extraction of domain-specific features plays an important role for image-to-image translation. There are three main challenges in solving the image-to-image translation problem. The first one is how to extract the domain-independent and domain-specific features for a given image. The second is how to merge the features from two different domains into a natural image in the target domain. The third one is that there is no paired data for us to learn such the mappings. In this paper, we focus on solving image-to-image translation problems with latent space disentanglement.

Although it is difficult to obtain the paired data across different image domains, we observe that in many cases, we are aware which domain an image comes from~\cite{choi2017stargan,liu2018unified}. Motivated by this observation, we propose to use this domain-level signal as explicit supervision and pre-train a deep convolutional neural network (CNN) to predict which domain an image is from. If such a network can well differentiate images from different domains, the output of the second-to-last layer of the network should carry rich domain information which captures each domain's specific characteristic. Therefore, we can leverage such a pre-trained CNN to extract the domain-specific features of an image. Compared with works \cite{lin2018conditional,huang2018multimodal,lee2018diverse} that use two separated domain-specific feature extractors for two domain translation, we utilize the domain classifier as a general domain-specific feature extractor which can be easily generalized to multi-domain translation. With the well-defined domain-specific features, the domain-independent features can be easily obtained by feature disentanglement.

Generally, in this work, we first propose a new frameworks for common image-to-image translation, which is composed by a domain-specific feature extractor, a domain-independent feature extractor, and an image generator. The domain-specific feature extractor is pre-trained based on domain supervision, and then the domain-independent feature extractor and the generator are trained based on Generative Adversarial Networks (GANs) and dual learning \cite{he2016dual}. We choose GAN and dual learning due to the following considerations: (1) The dual learning framework can help learn to extract and merge the domain-specific and domain-independent features by minimizing carefully designed reconstruction errors, including self-reconstruction errors and cross-domain reconstruction errors. (2) GAN can ensure that the generated images well mimic the natural images in the target domain. (3) Both dual learning \cite{he2016dual,Yi_2017_ICCV,zhu2017unpaired} and GAN \cite{goodfellow2014generative,radford2015unsupervised,denton2015deep} work well under unsupervised settings. Therefore, we name this framework as Domain-supervised GAN (briefly, DosGAN).

\begin{figure}[!t]
\centerline{\includegraphics[width=8.5cm]{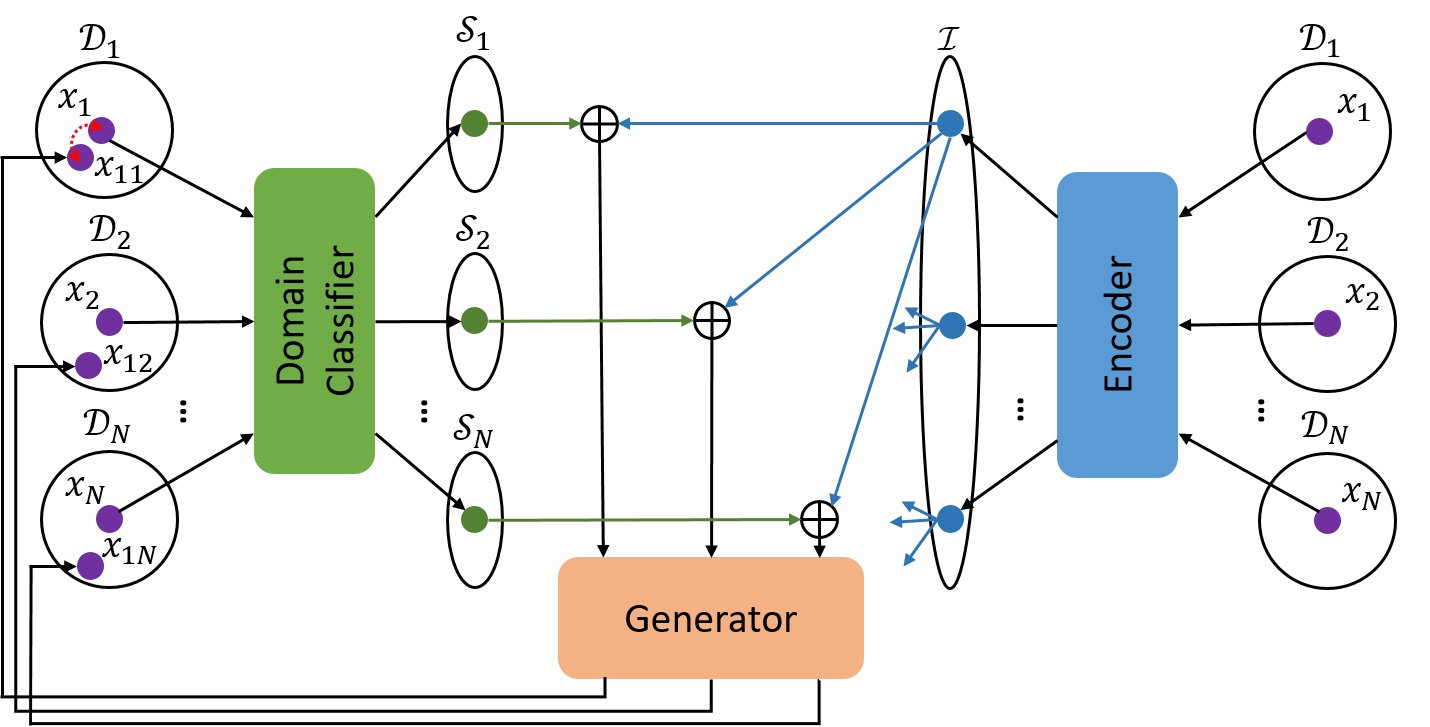}}
  \centering
\caption{The basic idea of our DosGAN-c on conditional image-to-image translation. $\mathcal{D}_k$ $\forall k\in[N]$ are $N$ domains of interests. Our model disentangles $N$ domain-specific feature latent space $\mathcal{S}_k$ through a classifier. Encoder in DosGAN-c learns to project images from $N$ domains to a domain-independent latent space $\mathcal{I}$ shared by all domains. Given $x_1\in\mathcal{D}_1$ for example, various translation image $x_{1k}$ can be obtained by varying domain-specific features in $\mathcal{S}_k$. Specially, $x_{11}$ should be consistent with $x_{1}$. }
\label{fig:basic_idea}
\end{figure}

Existing multi-domain translation models, such as StarGAN, also lack the ability to control the translated results in the target domain and their results usually lack of diversity in the sense that a fixed image usually leads to (almost) deterministic translation result. With two kinds of latent feature (i.e., domain-specific and domain-independent features) disentangled, we can devise a conditional DosGAN (briefly, DosGAN-c) for conditional image-to-image translation \cite{lin2018conditional}, which is to translate an image from the source domain to the target domain conditioned on a given image in the target domain and requires that the generated image should inherit some domain-specific features of the conditional image from the target domain. The DosGAN-c shares the same network architecture with DosGAN, and the difference between two frameworks only lies in the inputs and losses.

Compared with the previous work cd-GAN \cite{lin2018conditional}, there are three main differences: (1) Unlike cd-GAN or CycleGAN that simply treats multiple domains as different sources of images, in this work, we regard domain information as explicit supervision, where we train a classifier to classify the domain of  the input image, and convert it to a domain-specific feature extractor. (2) The domain-specific features in our model are explicitly modeled, which are extracted by a well-defined and fixed domain classification network. However, in cd-GAN, no such kind of explicit supervision signal is applied, causing the ambiguity in the extracted feature. (3) In this paper, we generalize the conditional image-to-image translation from two domains (with two models) to multiple domains ($\ge2$domains, with one model only), pushing the frontier of multi-domain conditional image-to-image translation.



We carry out experiments on multiple facial attribute translation, multiple season translation and conditional edges-to-shoes/handbags translation. The results demonstrate that with domain supervision, our approach can better extract domain-specific features and translate images.

Interestingly, we find that the pre-trained domain-specific feature extractor has certain transferability. We transfer the domain-specific feature extractor trained on the Facescrub dataset, where the domain of each image (i.e., the identity of each face image) is known, to the CelebA dataset, where the domain (i.e., identity) of each image is unknown. The two datasets share the same semantic representations (i.e., face images), which provides a viable way to produce the domain-specific features for the faces in the unknown domains.  Therefore, we can succeed in achieving conditional translation on CelebA by transferring the pre-trained domain-specific feature extractor, while previous methods cannot.

Our main contributions can be summarized as follows:
\begin{itemize}
    \item We explicitly exploit domain supervision in unpaired image-to-image translation by extracting well-defined domain-specific features through a simple domain classifier.
    \item We disentangle two latent features by explicitly utilizing domain supervision and achieve performances  better than several strong baseline methods.
    \item We show that the pre-trained domain-specific feature extractor can be transferred across datasets, which has not been investigated before.
\end{itemize}

The remaining parts of this paper are organized as follows. We review related work in Section \ref{sec:related work} and present our approach in Section \ref{sec:framework}.  The experimental results are reported in Section~\ref{sec:exps}. We conclude our work and discuss several future directions in the last section. Our code is publicly available to the research community at \url{https://github.com/linjx-ustc1106/DosGAN-PyTorch}.


\section{Related Works}\label{sec:related work}

\textbf{Generative modeling}. Image generation has been widely explored in recent years, and works have been putting effort on modeling the natural image distribution. This problem was initially solved by Boltzmann machines \cite{smolensky1986information} and autoencoders \cite{hinton2006reducing,vincent2008extracting}. Variational autoencoder (VAE) \cite{kingma2013auto} aims to improve the quality and efficiency of image generation by reparametrization of a latent distribution. GAN \cite{goodfellow2014generative} was firstly proposed to generate images from random variables by a two-player minimax game. Various works have been proposed to exploit the capability of GAN for various image generation tasks. InfoGAN \cite{chen2016infogan} learns to disentangle latent representations by maximizing the mutual information between a small subset of the latent variables and the observation. Radford et al. \cite{radford2015unsupervised} presented a class of deep convolutional generative networks (DCGANs) for high-quality image generation and unsupervised image classification tasks, which bridges the gap between convolutional neural networks (CNNs) and unsupervised image generation.

\textbf{Supervised image-to-image translation}. The supervised image-to-image translation aims to learn a parametric translation function that transforms an input image in a source domain to an image in a target domain when paired data is available. Many computer vision tasks can be posed as this problem. For example, Long et al. \cite{long2015fully} proposed a fully convolutional network (FCN) for image-to-segmentation translation. SRGAN \cite{ledig2016photo} maps low-resolution images to high resolution images. Isola et al. \cite{isola2016image} proposed a general conditional GAN (pix2pix) for image-to-image translation tasks, including label-to-street scene and aerial-to-map. In order to overcome the limitation of relatively low resolution in pix2pix and lack of realistic details and texture, Wang et al. \cite{2017arXiv171111585W} proposed a HD version of pix2pix which increases the resolution to 2048*1024. In this work, a coarse-to-fine generator, three multi-scale discriminators, and a feature matching loss are utilized. Like prior research \cite{tenenbaum2000separating} attempted to learn disentangled representations, such as content and style, Zhu et al. \cite{zhu2017toward}  proposed a BicycleGAN to learn multimodal image-to-image translation by building bijective consistency between output and latent code.

\textbf{Unpaired image-to-image translation}. Since it is usually expensive to collect a large amount of paired data for supervised image-to-image translation tasks, unpaired learning based algorithms have been widely adopted. Based on adversarial training, Dumoulin et al. \cite{dumoulin2016adversarially} and Donahue et al. \cite{donahue2016adversarial} proposed algorithms to jointly learn mappings between latent space and data bidirectionally. Taigman et al. \cite{taigman2016unsupervised} presented a domain transfer network (DTN) for unpaired cross-domain image generation by assuming constant latent space between two domains, which could generate images of target domain's style and preserve their identity. He et al. \cite{he2016dual} proposed a dual learning mechanism that can enable a neural machine translation system to automatically learn from unlabeled data through a dual learning game. Inspired by the idea of dual learning, DualGAN~\cite{Yi_2017_ICCV}, DiscoGAN~\cite{kim2017learning} and CycleGAN~\cite{zhu2017unpaired} were proposed to tackle the unpaired image translation problem by training two cross-domain transfer GANs at the same time. To generate controllable translation result, Lin et al. \cite{lin2018conditional} decompose the image latent space into domain-independent and domain-specific feature spaces, and raise a new problem named as conditional cross-domain translation which can assign domain-specific feature for generated result by feeding a conditional image in the target domain. Similar to \cite{lin2018conditional}, other two works \cite{huang2018multimodal,lee2018diverse} proposed to disentangle latent space and generate diverse translation results. Choi et al. \cite{choi2017stargan} further proposed a StarGAN that can perform image-to-image translations for multiple domains using only a single model. Similarly, Liu at al. \cite{liu2018unified} proposed a UFDN that learns domain-invariant representation from multiple domains and can perform continuous cross-domain image translation and manipulation. Some other works also utilized attention mechanism for more accurate image-to-image translation. DA-GAN \cite{Ma_2018_CVPR} firstly finds the instance-level corresponding of two domains in the latent space by introducing attention mechanism, then generates images from the highly-structured latent space. In \cite{pumarola2018ganimation}, GANimation is proposed to generate anatomically-aware facial animation. Attention mechanism is exploited to make the scheme more robust and have the capacity of dealing with images in the wild. Authors in \cite{NIPS2018_7627,chen2018attention} introduced attention map to avoid cycle consistency loss distracted by background factors and achieved better performance than original CycleGAN.


\section{Framework}\label{sec:framework}
In this section, we introduce our proposed framework. We first give a general formulation to image translation and conditional image translation, then describe our network architecture, and finally present the training algorithm and discussion.
\subsection{Problem formulation}\label{sec:problem}
We work on multiple domain translation with one single model, but to increase readability, we will introduce two domain translation between domain A (denoted as $\dom{A}$)  and domain B (denoted as $\dom{B}$). Suppose we have a domain-specific feature extractor $\alpha(\cdot)$ and a domain-independent feature extractor $\beta(\cdot)$.\footnote{We will discuss how to learn the two extractors in the following two subsections.} Given an image $x\in\mathcal{D}_A$ or $x\in\mathcal{D}_B$, we can get its domain-specific features $x^s$ and domain-independent features $x^i$ by applying the two extractors:
\begin{equation}
x^s=\alpha(x),\;x^i=\beta(x).
\end{equation}
In this paper, $x$ without any superscript denotes an image. $x$ with superscripts $i$ and $s$ refer to domain-independent features (e.g., $x^i$) and domain-specific features (e.g., $x^s$) respectively. Note that the domain-independent features of an image should be kept and its domain-specific features should be changed while translating it from one domain to other domain.

Then the style $S_A$ of $\dom{A}$ and the style $S_B$ of $\dom{B}$ are:
\begin{equation}
S_A = \int_{x\in\mathcal{D}_A}\alpha(x)p_A(x)\mathrm{d}x,\,S_B = \int_{x\in\mathcal{D}_B}\alpha(x)p_B(x)\mathrm{d}x,
\end{equation}
where $p_A(x)$ and $p_B(x)$ denote the probabilities of the image $x$ belonging to $\dom{A}$ and $\dom{B}$. Empirically, given a set $D_A$ of images in domain $A$ (i.e., $D_A\subset\dom{A}$) and a set of $D_B\subset\dom{B}$, $S_A$ and $S_B$ can be estimated through
\begin{equation}
S_A\approx\frac{1}{\vert D_A\vert}\sum_{x_A\in D_A}\alpha(x_A);\,S_B\approx\frac{1}{\vert D_B\vert}\sum_{x_B\in D_B}\alpha(x_B),
\label{eq:approx_domain_feature}
\end{equation}
where $|D_A|$ and $|D_B|$ refer to the number of images in $D_A$ and $D_B$.
Let $\oplus$ denote a generator, which takes a set of domain-specific features $x^s_\cdot$ and a set of domain-independent features $x^i_\cdot$ as inputs and generates an image in the corresponding domain: for any $x_A\in\dom{A}$ and $x_B\in\dom{B}$,
\begin{equation}
x_A=x^i_A\oplus x^s_A,\;x_B=x^i_B\oplus x^s_B.
\end{equation}
\noindent{\bf Definition 1} With the above notations, the translation from $\dom{A}$ to $\dom{B}$ and the translation from $\dom{B}$ to $\dom{A}$  can be written as follows: for any $x_A\in\dom{A}$ and $x_B\in\dom{B}$,
\begin{equation}
\begin{aligned}
x_{AB}=x^i_A\oplus S_B,\;x_{BA}=x^i_B\oplus S_A,\;
\end{aligned}
\label{eq:non_conditional}
\end{equation}
where $x_{AB}$ and $x_{BA}$ denote the translation result.

\noindent{\bf Definition 2} The conditional image-to-image translation of an image $x_A$ in domain $A$ conditioned on an image $x_B$ in domain $B$, i.e., $\dom{A}\times\dom{B}\mapsto\dom{B}$, and the dual translation $\dom{B}\times\dom{A}\mapsto\dom{A}$, can be written as
\begin{equation}
\begin{aligned}
x_{AB}=x^i_A\oplus x^s_B,\;x_{BA}=x^i_B\oplus x^s_A.
\end{aligned}
\label{eq:conditional}
\end{equation}
Different from existing works~\cite{lin2018conditional,huang2018multimodal,lee2018diverse}, in this work, the domain-specific feature extractor $\alpha(\cdot)$ is pre-trained in a supervised setting based on domain supervision and will be kept fixed during the training of the translation system. In the following subsections, we will first discuss how to pre-train $\alpha(\cdot)$, and then present how to learn the domain-independent feature extractor $\beta(\cdot)$ and the generator $\oplus$ from unpaired images for image-to-image translation and conditional image-to-image translation.

\subsection{Pre-training of domain-specific feature extractor}
Very different from previous works~\cite{zhu2017unpaired,lin2018conditional}, which simply treat multiple domains as different sources of images, in this work, we regard them as explicit supervision and use them in a supervised way to learn the domain-specific feature extractor $\alpha(\cdot)$.

Given images from $N$ domains, we train a CNN network to correctly classify the domain of an image. After training, if the classifier network can accurately detect the domain of an image, then the output of the second-to-last layer of this classifier should well capture the domain information of this image. Thus, we denote the input layer to the second-to-last layer of this classifier network as our domain-specific feature extractor $\alpha(\cdot)$. With this pre-trained $\alpha(\cdot)$, we will introduce our system architectures for image translation in the coming subsections.

\subsection{Architecture for image-to-image translation}
\begin{figure*}[!t]
  \centerline{\includegraphics[width=14.5cm]{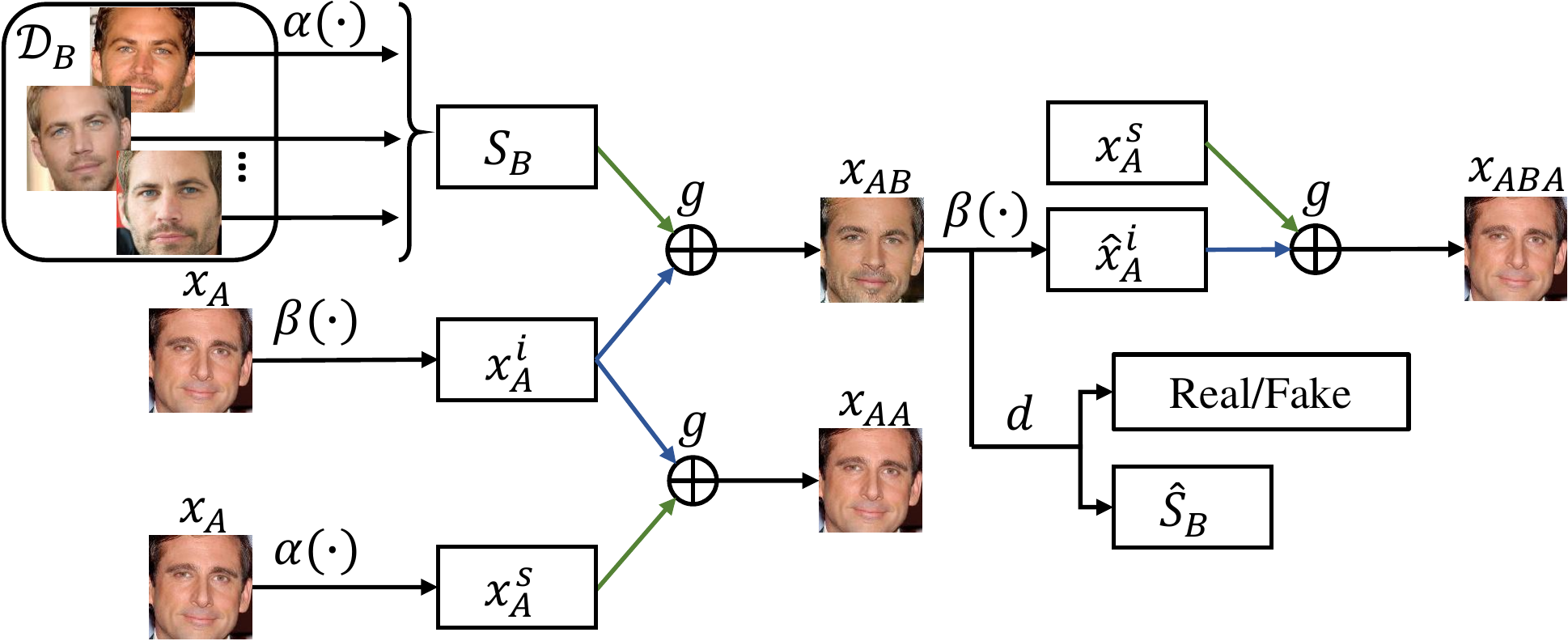}}
  \caption{Training architecture of the proposed DosGAN for unpaired image-to-image translation.}
  \centering
\label{fig:nonconditional_framework}
\end{figure*}
The architecture of our proposed system for image-to-image translation is shown in Figure~\ref{fig:nonconditional_framework}. As can be seen, to translate an image $x_A$ from domain $A$ to domain $B$, we first apply the domain-specific feature extractor $\alpha(\cdot)$ over all the images in domain $B$ and obtain the style feature $S_B$ according to Eqn~\eqref{eq:approx_domain_feature}. Then we apply the  domain-independent feature extractor $\beta(\cdot)$ to $x_A$ and obtain the domain-independent features $x^i_A$. After that, the generator $\oplus$ takes $x^i_A$ and $S_B$ as inputs and generates the translation results $x_{AB}$, i.e., $x_{AB}=x^i_A\oplus S_B$. In practice, the generator is modeled by a neural network and needs to be learned, thus, to be differentiated from the oracle generator $\oplus$, we use $g$ to denote the generator to be learned.

\noindent{\bf Adversarial loss} Following the idea of GANs, we introduce a discriminator $d$, which takes a (real or fake) image as input and output a probability indicating how likely the input belongs to a real image domain. We illustrate the objective function as below:
\begin{equation}
\begin{aligned}
\ell_{\text{GAN}}  = \frac{1}{\vert D_A\vert}\sum_{x_{A}\in D_A}&\log(d_{\text{adv}}(x_A))
+\log(1-d_{\text{adv}}(x_{AB})),
\end{aligned}
\label{eq:gan_loss}
\end{equation}
where $x_{AB}$ is defined in Eqn.~\eqref{eq:non_conditional}. The $g\circ \beta$ network tries to fool the discriminators by minimizing $\ell_{\text{GAN}}$, while the discriminator tries to differentiate generated images from real ones by maximizing $\ell_{\text{GAN}}$.

\noindent{\bf Domain-specific reconstruction loss}  Different from StarGAN~\cite{choi2017stargan} that requires discriminator $d$ to serve as a classifier that distinguishes which domain the input belongs to, we adapt the discriminator $d$ to maximize the mutual information between domain-specific features and generated images as InfoGAN~\cite{chen2016infogan}. In this case, the domain-specific feature reconstruction cost $\ell_{\alpha,r}$ of real images and fake images $\ell_{\alpha,f}$ can be formulated as follows:
\begin{equation}
\begin{aligned}
\ell_{\alpha,r}&=\frac{1}{\vert D_A\vert}\sum_{x_{A}\in D_A}[\Vert d_{\text{feat}}(x_{A}) - x_A^s \Vert_1],\\
\ell_{\alpha,f}& =\frac{1}{\vert D_A\vert}\sum_{x_{A}\in D_A}[\Vert d_{\text{feat}}(x_{AB}) - S_B \Vert_1],
\end{aligned}
\label{eq:ds_reconstruct_loss}
\end{equation}
where $d_{\text{feat}}(x)$ represents the domain-specific features predicted from input $x$ by discriminator $d$. The discriminator $d$ minimizes the $\ell_{\alpha,r}$ to learn to reconstruct correct domain-specific features from real images, and the $g\circ \beta$ network minimizes the $\ell_{\alpha,f}$ to generate image containing correct style feature in the target domain.

\noindent{\bf Image reconstruction loss} We use two kinds of reconstruction losses here. (1) Self-reconstruction loss, which is to minimize the L1 norm between $x_A$ and $x_{AA}=g(x^i_A,x_A^s)$. With such a loss, after given the domain-specific features $x_A^s$ extracted by $\alpha(\cdot)$, we can ensure what $\beta(\cdot)$ extracts is the domain-independent features.(2) Cross-domain loss, which is to minimize the L1 norm between $x_A$ and $x_{ABA}=g(\beta(x_{AB}),x_A^s)$. Thus, the image level reconstruction loss is:
\begin{equation}
	\begin{aligned}
		\ell_{\text{im}}= \frac{1}{\vert D_A\vert}\sum_{x_{A}\in D_A}[\Vert x_A - x_{AA} \Vert_1+\Vert x_A - x_{ABA} \Vert_1].
	\end{aligned}
	\label{eq:reconstruct_images}
\end{equation}
\noindent{\bf Overall training loss}
We combine the above three losses to optimize the framework. To optimize the domain-independent feature extractor $\beta(\cdot)$ and generator $g$ (remind that $\alpha$ is fixed), we will minimize
\begin{equation}
\begin{aligned}
\ell^{\text{total}}_{\text{net}}= \ell_{\text{GAN}}+\lambda_{f}\ell_{\alpha,f}+\lambda_{\text{im}}\ell_{\text{im}}.
\label{eq:total_loss_generator}
\end{aligned}
\end{equation}
To optimize the discriminator, we need to minimize
\begin{equation}
\begin{aligned}
\ell^{\text{total}}_{d}= - \ell_{\text{GAN}} + \lambda_{f}\ell_{\alpha,r},
\label{eq:total_loss_gan}
\end{aligned}
\end{equation}
where $\lambda_{f}$ and $\lambda_{\text{im}}$ are weights to achieve balance among different loss terms. 

\subsection{Architecture for conditional image-to-image translation}
\begin{figure*}[!t]
  \centerline{\includegraphics[width=13.5cm]{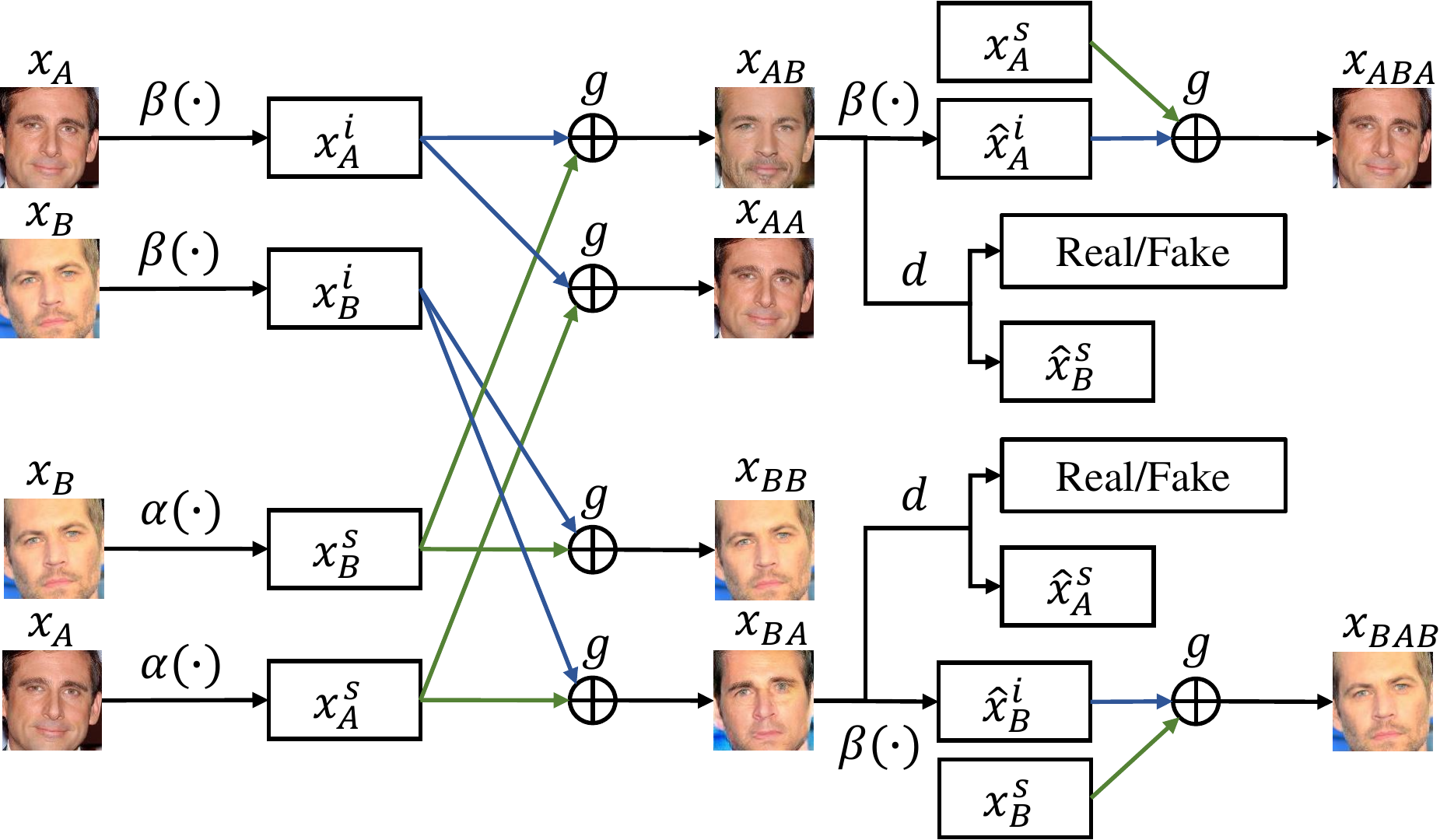}}
  \caption{Training architecture of the proposed DosGAN for unpaired conditional image-to-image translation (briefly, DosGAN-c).}
  \centering
\label{fig:conditional_framework}
\end{figure*}
The architecture of DosGAN for conditional image-to-image translation (DosGAN-c) is shown in Figure \ref{fig:conditional_framework}. As can be seen, to translate an image $x_A$ from domain $A$ to domain $B$ conditioned on an image $x_B$ from domain $B$, we first apply the domain-specific feature extractor $\alpha(\cdot)$ on $x_B$ and obtain the domain-specific feature $x^s_B$. Then we apply the  domain-independent feature extractor $\beta(\cdot)$ to $x_A$ and obtain the domain-independent features $x^i_A$. After that, the generator $g$ takes $x^i_A$ and $x^s_B$ as inputs and generates the translation results $x_{AB}$. Similarly, $x_{BA}$ is also generated. A mathematical definition is in Definition 2. In this setting, for ease of reference, we set $|D_A|=|D_B|$.

\noindent{\bf Adversarial loss} Similar to non-conditional setting, we illustrate the adversarial loss as below:
\begin{equation}
\begin{aligned}
\ell^c_{\text{GAN}} =& \frac{1}{\vert D_A\vert}\sum_{x_{A}\in D_A}&\log(d_{\text{adv}}(x_A))
+\log(1-d_{\text{adv}}(x_{AB}))\\+& \frac{1}{\vert D_B\vert}\sum_{x_{B}\in D_B}&\log(d_{\text{adv}}(x_B))
+\log(1-d_{\text{adv}}(x_{BA})).
\end{aligned}
\label{eq:gan_loss_condition}
\end{equation}
\noindent{\bf Domain-specific reconstruction loss}  Different from non-conditional setting, the domain-specific feature reconstruction cost $\ell_{\alpha,r}$ of real images and $\ell_{\alpha,f}$ of fake images  are modified as follows:
\begin{equation}
\begin{aligned}
\ell^c_{\alpha,r}=\frac{1}{\vert D_A\vert}\sum_{x_{A}\in D_A}[&\Vert d_{\text{feat}}(x_{A}) - x_A^s \Vert_1]
\\+\frac{1}{\vert D_B \vert}\sum_{x_{B}\in D_B}[&\Vert d_{\text{feat}}(x_{B}) - x_B^s \Vert_1],\\
\ell^c_{\alpha,f}=\frac{1}{\vert D_B\vert}\sum_{\substack{x_{A}\in D_A \\ x_{B}\in D_B} }[&\Vert d_{\text{feat}}(x_{AB}) - x_B^s \Vert_1 \\
+& \Vert d_{\text{feat}}(x_{BA}) - x_A^s \Vert_1].
\end{aligned}
\label{eq:cls_loss_1}
\end{equation}
The discriminator $d$ minimizes the $\ell^c_{\alpha,r}$ to learn to reconstruct correct domain-specific features from real images, and the $g\circ \beta$ network minimizes the $\ell^c_{\alpha,f}$ to generate image containing correct domain-specific features from image in the target domain.

\noindent{\bf Image reconstruction loss} Similar to the non-conditional image-to-image translation, the reconstructed images are defined as follows:
\begin{equation}
\begin{aligned}
& x_{AA}=g(x_A^i,x_A^s);\;x_{BB}=g(x_B^i,x_B^s);\\
& x_{ABA}=g(\beta(x_{AB}),x_A^s);\;x_{BAB}=g(\beta(x_{BA}),x_B^s).
\end{aligned}
\end{equation}
Correspondingly, the reconstruction loss is
\begin{equation}
\begin{aligned}
\ell^c_{\text{im}}= \frac{1}{\vert D_A\vert}\sum_{\substack{x_{A}\in D_A\\x_{B}\in D_B} }\Big[&\Vert x_A - x_{AA} \Vert_1+\Vert x_A - x_{ABA} \Vert_1\\
+&\Vert x_B - x_{BB} \Vert_1+\Vert x_B - x_{BAB} \Vert_1
\Big].
\end{aligned}
\label{eq:dual_im}
\end{equation}
\noindent{\bf Overall training loss}
The total training loss of the generation network and discriminator are defined as follows:
\begin{equation}
\begin{aligned}
 \ell^{\text{total,c}}_{\text{net}}= \ell^c_{\text{GAN}}+\lambda_{f}\ell^c_{\alpha,f}+\lambda_{\text{im}}\ell^c_{\text{im}},
\label{eq:total_loss_generator_c}
\end{aligned}
\end{equation}

\begin{equation}
\begin{aligned}
 \ell^{\text{total,c}}_{d}= - \ell^c_{\text{GAN}} + \lambda_{f}\ell^c_{\alpha,r},
\label{eq:total_loss_gan_c}
\end{aligned}
\end{equation}

where $\lambda_{f}$ and $\lambda_{\text{im}}$ are weights to achieve balance among different loss terms. The $g\circ \beta$ and $d$ should try to minimize $\ell^{\text{total,c}}_{\text{net}}$ and $\ell^{\text{total,c}}_{d}$ respectively. The overall process is summarized in Algorithm~\ref{alg:intnet}.
\begin{algorithm}[!htpb]
	\caption{DosGAN/DosGAN-c training process.}
	\label{alg:intnet}
	\begin{algorithmic}[1]
		\State{\em Input:} $N$ image domains $\mathcal{D}_k$ $\forall k\in[N]$, batch size $K$, learning rate $\eta$;
		\State Randomly initialize the parameters $\Theta_\alpha$ of domain-specific feature extractor $\alpha(\cdot)$.
        \State Randomly select one domain $\mathcal{D}_k$, $k\in[N]$. Get a minibatch of data $D_k$ satisfying $D_k\subset\mathcal{D}_k$ and $\vert D_k\vert=K$;
        \State Update the classifier as follows:\newline
        $\Theta_\alpha \leftarrow \Theta_\alpha - \eta\nabla_{\Theta_\alpha}\frac{-1}{\vert D_k\vert}\sum_{x\in D_k}\log P(k|x;\Theta_\alpha)$;
        \State Repeat step 3 and step 4 until convergence.
        \State Randomly initialize the parameters $\Theta_{g\circ \beta}$ of $g$ and $\beta(\cdot)$, and parameters $\Theta_d$ of discriminator $d$;
		\State Randomly select two different domains $\mathcal{D}_A$, $\mathcal{D}_B$, $A,B\in[N]$. For each selected domain $\mathcal{D}_l$ where $l\in\{A,B\}$, get a minibatch of data $D_l$ satisfying $D_l\subset\mathcal{D}_l$ and $\vert D_l\vert=K$.
        \If{Training DosGAN}
		\State Update the parameters as follows:
		\begin{equation}
		\begin{aligned}
		& \Theta_{g\circ \beta} \leftarrow \Theta_{g\circ \beta} - \eta\nabla_{\Theta_{g\circ \beta}}\ell^{\text{total}}_{\text{net}}(D_A), \\
		& \Theta_d \leftarrow \Theta_d - \eta\nabla_{\Theta_d}\ell^{\text{total}}_{d}(D_A),
		\end{aligned}
		\end{equation}
		\State where $\ell^{\text{total}}_{\text{net}}(D_A)$ and $\ell^{\text{total}}_{d}(D_A)$ are defined in Eqn.~\eqref{eq:total_loss_generator} and Eqn.~\eqref{eq:total_loss_gan} respectively.
        \EndIf
        \If{Training DosGAN-c}
        \State Update the parameters as follows:
		\begin{equation}
		\begin{aligned}
		& \Theta_{g\circ \beta} \leftarrow \Theta_{g\circ \beta} - \eta\nabla_{\Theta_{g\circ \beta}}\ell^{\text{total,c}}_{\text{net}}(D_A,D_B), \\
		& \Theta_d \leftarrow \Theta_d - \eta\nabla_{\Theta_d}\ell^{\text{total,c}}_{d}(D_A,D_B),
		\end{aligned}
		\end{equation}
		\State where $\ell^{\text{total,c}}_{\text{net}}(D_A,D_B)$ and $\ell^{\text{total,c}}_{d}(D_A,D_B)$ are defined in Eqn.~\eqref{eq:total_loss_generator_c} and Eqn.~\eqref{eq:total_loss_gan_c} respectively.
        \EndIf
		\State Repeat step 7 and step 13 until convergence.

	\end{algorithmic}
\end{algorithm}
\subsection{Discussion}\label{sec:discussion}
Our proposed framework can learn to extract the domain-independent features and combine them with domain-specific features in one single encoder-decoder network. Take the conditional DosGAN as an example. Consider the path of $x_A\rightarrow \beta(\cdot)\rightarrow (x_A^i, x_A^s)\rightarrow g\rightarrow x_{AA}$ and path of $x_A\rightarrow \beta(\cdot)\rightarrow (x_A^i, x_B^s)\rightarrow g\rightarrow x_{AB}$. Our training objective requires that $x_{AA}$ is an image in domain $\mathcal{D}_A$ and $x_{AB}$ is an image in domain $\mathcal{D}_B$. Given the fact that differences between $x_{AA}$ and $x_{AB}$ lay in the differences between  $x_A^s$ and $x_B^s$, thus it is implied that  $x_A^s$ and $x_B^s$ are domain-specific features and are extracted by a pre-trained domain-specific feature extractor $\alpha(\cdot)$.  Compared with \cite{lin2018conditional,huang2018multimodal,lee2018diverse},  the domain-specific features in our model are extracted by a well-defined and fixed domain classification network, which avoids the feature ambiguity when both domain-specific feature and domain-independent feature extractors are trained together. With domain-specific features being given, $x_A^i$ should try to inherent the features in $x_{AA}$ and $x_{AB}$, which is further constrained with reconstruction loss $\ell^c_{\text{im}}$. Compared with StarGAN \cite{choi2017stargan} that is trained to reconstruct the source input with source domain code, which may cause a semi-optimum when images in one domain vary their own characteristic very much, such as the same identity may have different face appearance at different time, our model is trained with features from real images in every training step, and ideally can accurately reconstruct source input with its own domain-specific features.

\section{Experiments}\label{sec:exps}
\subsection{Implementation details}
\begin{figure}[!htbp]
	\centering
	\subfigure[$\beta(\cdot)$ Configuration]{
		\includegraphics[height=0.5\linewidth]{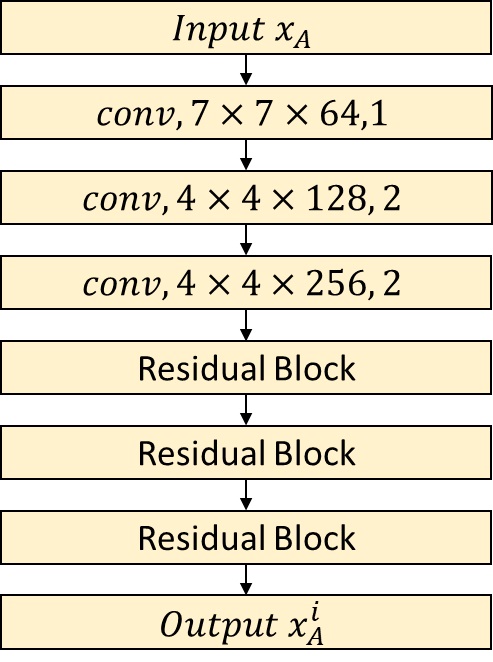}	
		\label{fig:enc_struc}
	}
	\subfigure[Generator Configuration]{
		\includegraphics[height=0.5\linewidth]{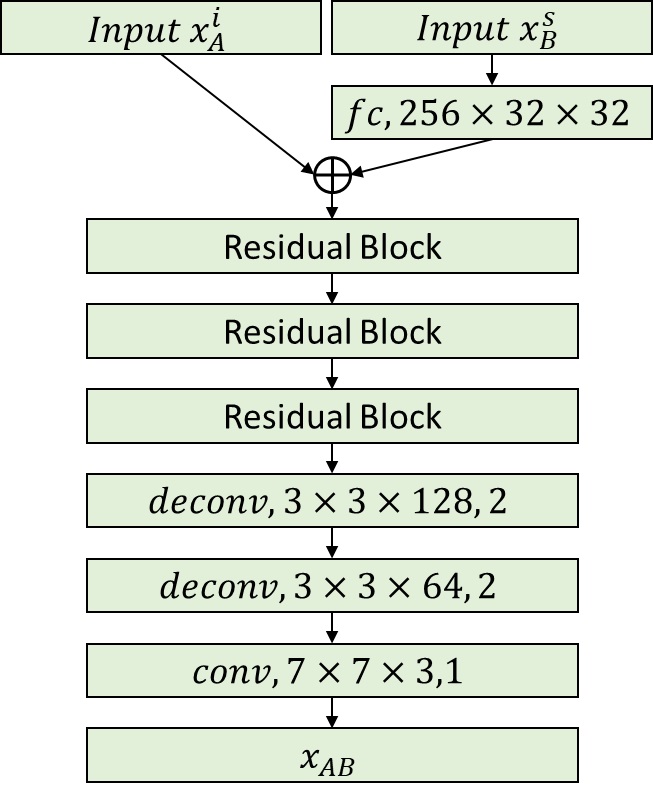}	
		\label{fig:dec_struc}
	}
	\subfigure[Discriminator Configuration]{
		\includegraphics[height=0.5\linewidth]{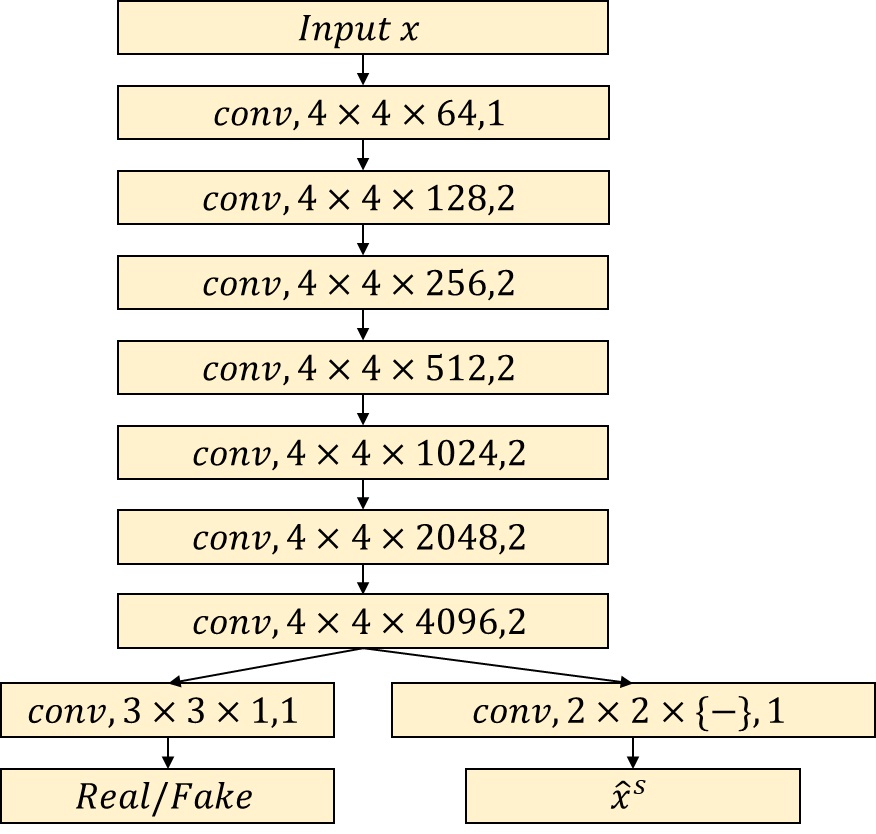}	
		\label{fig:D_struc}
	}
    \subfigure[$\alpha(\cdot)$ Configuration]{
		\includegraphics[height=0.5\linewidth]{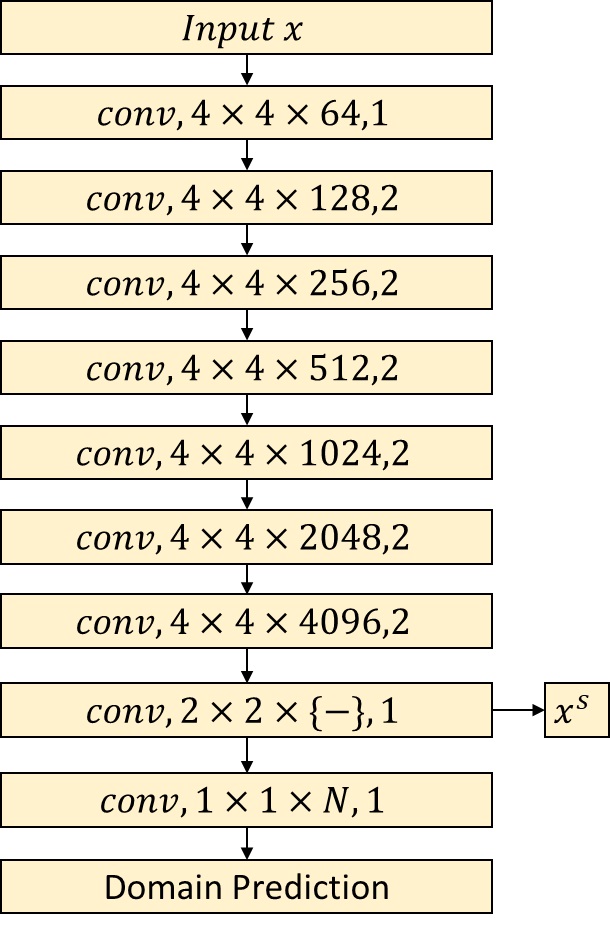}	
		\label{fig:C_struc}
	}
	\caption{The network configuration of DosGAN/DosGAN-c. $\{-\}$ represents the number of domain-specific features which is determined by the different tasks. }
	\label{fig:system_struc}
\end{figure}

In the following experiments, our DosGAN/DosGAN-c's network configuration is shown in Figure \ref{fig:system_struc}. For $\alpha(\cdot)$, it consists of one convolution layer with stride $1$ and kernel size $4\times 4$; six convolution layers with stride $2$ and kernel size $4\times 4$; the last two convolution layers that are implemented for domain-specific feature output and classification output. The number of domain-specific features is set to $1024$ for multiple identity translation and multiple facial attribute translation, $64$ for multiple season translation, $16$ for conditional cross-domain translation according to different tasks' domain classification accuracy and feature compactness. For discriminator $d$, we use PatchGANs \cite{isola2016image} that consists of six convolution layers with stride $2$ and kernel size $4\times 4$, and two separated convolution layers that are implemented for discrimination output $d_{\text{adv}}(x)$ and domain-specific feature reconstruction output $d_{\text{feat}}(x)$. For $\beta(\cdot)$, it one convolution layer with stride $1$ and kernel size $7\times 7$, two convolution layers with stride $2$ and kernel size $4\times 4$, and $3$ residual blocks \cite{he2016deep}. Each convolution layer is followed by Instance Normalization (IN) \cite{Ulyanov2016Instance} and ReLU units \cite{nair2010rectified}. For generator $g$, it first processes the domain-specific features through a fully-connected layer and adds it to domain-independent features from encoder $e$. Then the combined feature is input to $3$ residual blocks, two deconvolution layers with stride $2$ and kernel size $3\times 3$ followed by IN and ReLU units, and one convolution layer with stride $1$ and kernel size $7\times 7$.

We set the parameters $\lambda_{f}=5$ for multiple identity translation and multiple facial attribute translation, $\lambda_{f}=0.15$ for multiple season translation, $\lambda_{f}=0.5$ for conditional cross-domain translation, and $\lambda_{\text{im}}=10$ for all the experiments. We train our networks using Adam \cite{kingma2014adam} with learning rate of $0.0001$. For all experiments, we train models with a learning rate of $0.0001$ in the first $100000$ iterations and linearly decay the learning every $1000$ iteration.

\subsection{Baselines}
To verify the generality of our model, we compare with four state-of-the-art image translation models for different image-to-image translation tasks.

\textbf{CycleGAN}~\cite{Yi_2017_ICCV,kim2017learning,zhu2017unpaired}. CycleGAN was proposed for cross-domain translation task, which jointly trains two mappings with adversarial loss and dual loss.

\textbf{BicycleGAN}~\cite{zhu2017toward}. BicycleGAN was proposed for multimodal cross-domain translation and learns to combine input image with latent code. We compare with BicycleGAN when the paired data is available.

\textbf{cd-GAN}~\cite{lin2018conditional}. cd-GAN was proposed for unpaired conditional cross-domain translation, which aims to translate an image from source domain conditioned on a given image in the target domain. We show that our model can also deal with this problem.

\textbf{StarGAN}~\cite{choi2017stargan}. StarGAN can perform multi-domain translation using a single model. We compare with StarGAN in the case of multi-domain translation.

\textbf{MUNIT} \cite{huang2018multimodal} and \textbf{DRIT} \cite{lee2018diverse}. Both works are proposed for unpaired multimodal cross-domain translation, and learn to disentangle latent space and generate diverse translation results.
\subsection{Datasets}
For multi-domain translation, we compare our model with StarGAN on multiple facial attribute (three hair colors, including black hair, blond hair, brown hair; two genders, inlcuding male and female; two ages, including old and young) translation on CelebA dataset~\cite{liu2015faceattributes}, multiple season translation and multiple identity translation on both Facescrub dataset \cite{ng2014data} and CelebA.

For multiple facial attribute translation, the initial images in CelebA whose sizes are $178\times 218$ are resized to $128\times 156$. The test set is built by randomly selecting $2000$ images from the original dataset. All the remaining images are used as the training set.

For multiple season translation, the season dataset \cite{anoosheh2018combogan} consists of approximately $6000$ images and are categorized
into four seasons, i.e.,  Spring, Summer, Autumn and Winter. All images are resized to $256\times 256$.

For multiple identity translation, the Facescrub dataset \cite{ng2014data} comprises more than $100$ thousands face images of $531$ male and female celebrities, with about $200$ images per person. We resize all face images to $128\times 128$ and obtain $531$ identity domains. We build the test set by randomly selecting $20$ images per person from the original dataset. All the remaining images are used as the training set. The face images of CelebA dataset contains $10177$ identities but do not have corresponding identity labels. We apply a cropping box $(25,60,133,168)$ on CelebA images and resize them to $128\times 128$.

For conditional cross-domain translation \cite{lin2018conditional}, we carry out experiments on edges$\rightarrow $shoes dataset \cite{yu2014fine} and edges$\rightarrow $handbags dataset \cite{zhu2016generative}.

\begin{figure*}[htb!]
	\centering
	\centerline{\includegraphics[width=18.5cm]{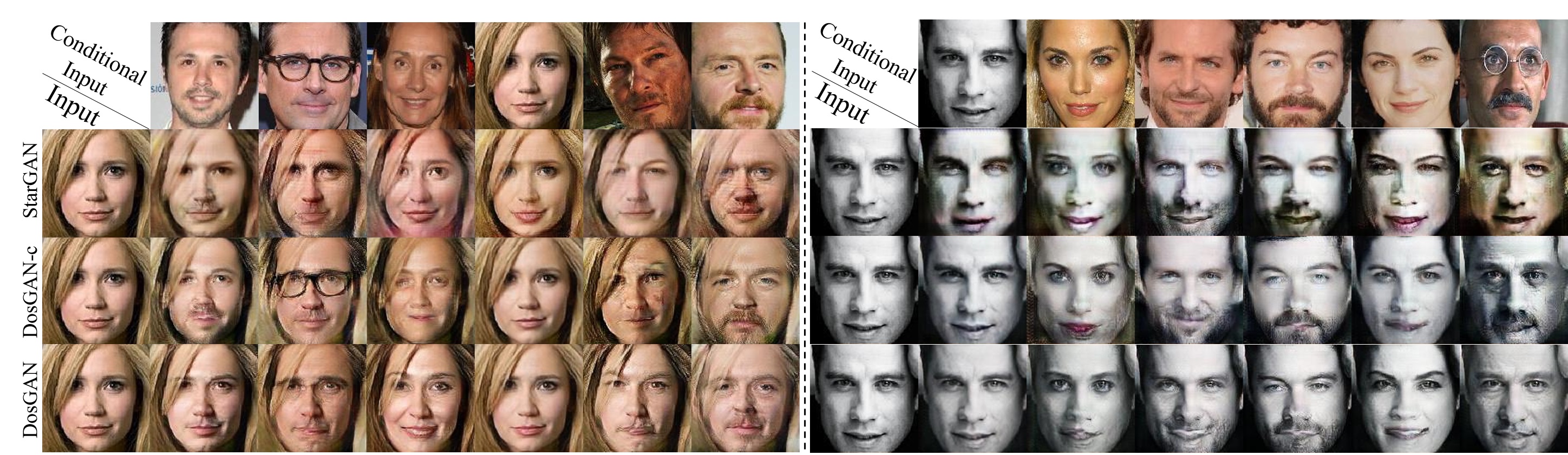}}
	\caption{Multiple identity translation results of StarGAN and our DosGAN on Facescrub. The first column is inputs $x_A$, and the first row is $x_B$ for DosGAN-c's conditional inputs and target identity examples for StarGAN and DosGAN. Other images are the translation results.}
	\label{fig:facescrub_intra}
\end{figure*}
\subsection{Multiple identity translation}\label{sec:intra_domain_trans}
We conduct multiple identity translation with DosGAN and conditional DosGAN (briefly, DosGAN-c) as illustrated in Eqn.~\eqref{eq:non_conditional} and Eqn.~\eqref{eq:conditional} respectively. Given random inputs and target identities in conditional inputs, the results of multiple identity translation compared with StarGAN~\cite{choi2017stargan} are shown in Figure~\ref{fig:facescrub_intra}. In general, our proposed method can successfully transfer the identity of input image to identity of the conditional input and greatly outperforms StarGAN. There are several aspects of observation.

\noindent(1) Results of our DosGAN are much visually better than StarGAN's. One main reason may be that using manually-set domain code without semantic meaning for domain representation is inappropriate and insufficient for multiple domain translation, especially when the domain number is large (i.e., $N=531$). On the contrary, our results are generated by combing domain-independent features and domain-specific features which are well disentangled and are both from real image inputs. Therefore, our model does not need to learn to relate domain code with corresponding identity of each image, but only needs to focus on generating results with two kinds of well-defined features.

\noindent(2) Our DosGAN-c can translate not only the input to the target identity, but also the specific face details (such as makeup, eyeglass) in the conditional input, while StarGAN can only translate input to the deterministic output without specific details.

\noindent(3) Both DosGAN and DosGAN-c can more accurately reconstruct the original inputs $x_A$ than StarGAN as the images at the fifth column in the left figure and the first column in the right figure. This is because StarGAN is required to reconstruct the original image with original domain code, but domain code for the same identity may have different face appearance at different time, which causes a relatively hard optimization for reconstruction accuracy. Our domain-specific features can effectively avoid this problem since the reconstruction of $x_A$ is generated by domain-specific features from itself.

\begin{table}[tb!]
\footnotesize
\setlength{\abovecaptionskip}{0.cm}
\setlength{\belowcaptionskip}{-0.cm}
\centering
\caption{Top-1 and top-5 face recognition accuracy [\%] of multiple identity translation results on Facescrub.}
\label{table:face_reg_acc_1}
\scalebox{1.0}{
\begin{tabular}{c c c c c}
\toprule
    &   StarGAN \cite{choi2017stargan}    &DosGAN-c &DosGAN &Real Faces \\
\midrule
Top-1 & 55.04 &60.79 & 94.01    & 80.20\\
Top-5 & 71.34 & 75.92 & 97.98 &95.13\\
\bottomrule
\end{tabular}}
\end{table}

For quantitative evaluation, we argue that if images translated by a model are accurately classified as the target identity's classes, the model is successful on multiple identity translation. We translate source image to random target identity in each test minibatch, and report the top-1 and top-5 face recognition accuracy of generated images from StarGAN and our model in Table~\ref{table:face_reg_acc_1} with re-trained $\text{VGG-16}$ network \cite{simonyan2014very}. We can see that our proposed method achieves better face recognition accuracy than StarGAN in both settings.

We also surprisingly find that the classification accuracy of images generated by DosGAN is much higher than that of real faces in the test set. To explain this phenomenon, we split the training set into two parts: One is used to train the classifier $\text{VGG-16}$ (briefly denoted as $\text{VGG-16}_{\text{half}}$), and the other is used to train DosGAN (briefly denoted as DosGAN$_{\text{half}}$). The face recognition accuracy of two VGG-16s, that are trained on full training set and half training set respectively, on training images, testing images, generated images from two DosGANs is shown in Table \ref{table:face_reg_acc_split}. We can find that when $\text{VGG-16}$ and DosGAN are trained with different training sets, classification accuracy of generated images from DosGAN$_{\text{half}}$ (80.98$\%$)  is only slightly higher than testing images's (78.42$\%$). This is because our model can effectively make use of training images' domain-specific features and carry the learned features for image generation in the testing phase (we take the expectation over domain-specific features from all training images as the domain characteristics). Therefore, when DosGAN and $\text{VGG-16}$ are trained with same dataset, DosGAN will generate images that have similar characteristics like training images and cause a similar classification accuracy as training images. Even for $\text{VGG-16}_{\text{half}}$, our model has higher accuracy than test images since a generated image containing the expectation of domain features is more capable to present the special characteristics of an identity than a single image.


\begin{table}[tb!]
\footnotesize
\setlength{\abovecaptionskip}{0.cm}
\setlength{\belowcaptionskip}{-0.cm}
\centering
\caption{Top-1 face recognition accuracy [\%] of two VGG-16s, that are trained on full training set and half training set respectively, on training images, testing images, generated images from DosGAN and generated images from DosGAN$_{\text{half}}$.}
\label{table:face_reg_acc_split}
\scalebox{1.0}{
\begin{tabular}{c c c c c c}
\toprule
&Training Faces &Testing Faces &DosGAN & DosGAN$_{\text{half}}$   \\
\midrule
$\text{VGG-16}$ & 99.68  & 80.20    & 94.01 & $-$\\
$\text{VGG-16}_{\text{half}}$ & 99.53 & 78.42 & $-$ & 80.98\\
\bottomrule
\end{tabular}}
\end{table}

\subsection{Adapting translation from Facescrub to CelebA}\label{sec:trans_adapt¡ª_tune}
\begin{figure*}[tb!]
	\centering
	\centerline{\includegraphics[width=18.5cm]{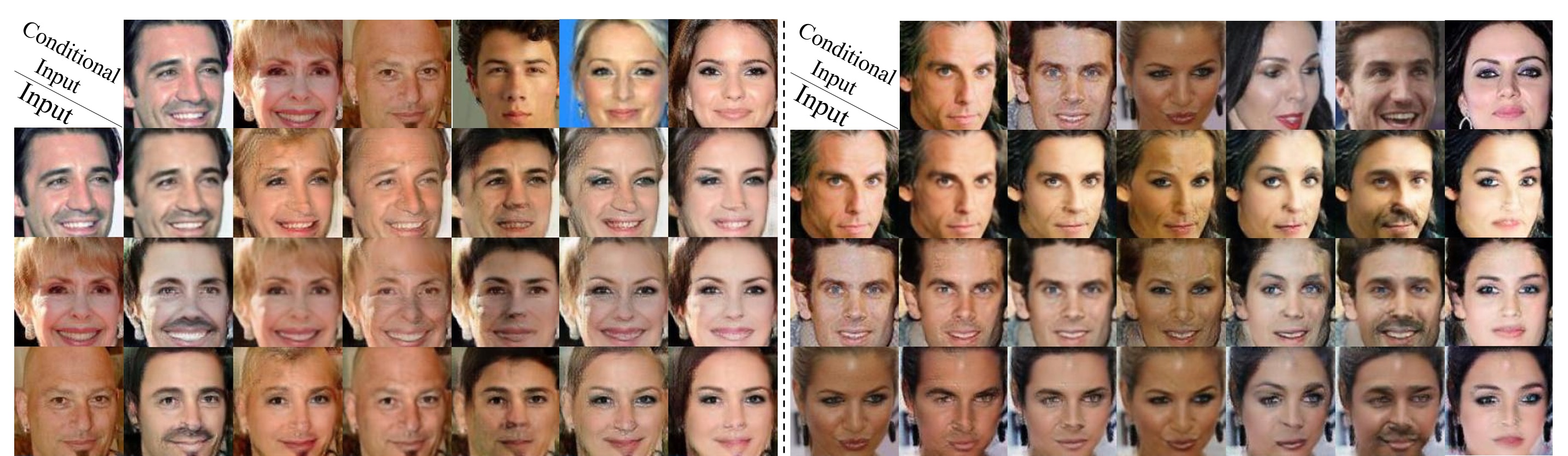}}
	\caption{Our results on multiple identity translation on unlabeled CelebA with domain supervision from Facescrub.}
	\label{fig:celeba_intra}
\end{figure*}
\begin{figure*}[tb!]
	\centering
	\centerline{\includegraphics[width=18.5cm]{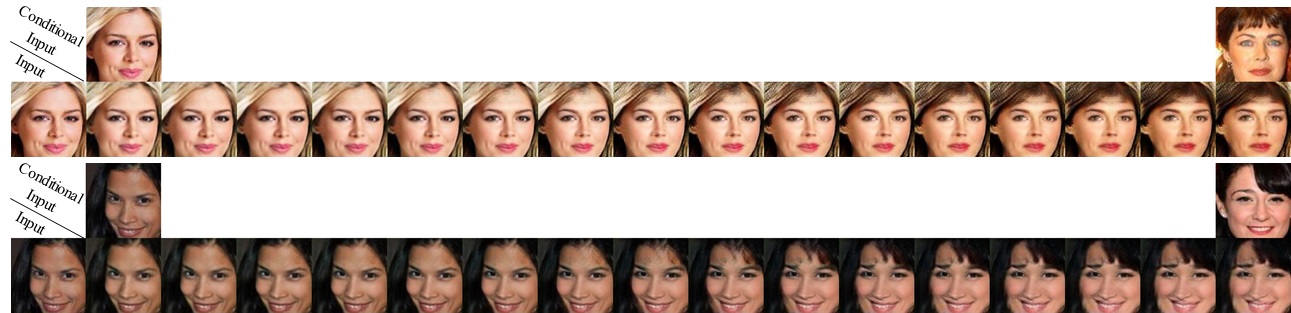}}
	\caption{Domain-specific interpolations on the CelebA. The translation results are generated by combining input's domain-independent features with domain-specific features interpolated linearly from left conditional input's  to right conditional input's. }
	\label{fig:celeba_intra_interp}
\end{figure*}
Most existing image-to-image translation works focus on translation within dataset that has clear domain information. However, there are still many datasets that does not contain domain information. An example is CelebA: for any photo in CelebA, we do not know which identity this photo belongs to. As far as we know, no previous work has touched how to achieve translation on dataset without domain separation, since there is no way to extract domain-specific features.

Considering both CelebA and Facescrub are face datasets, we can transfer the $\alpha(\cdot)$ trained on Facescrub to that for CelebA, and then retrain the $\beta(\cdot)$ and $g$ in the DosGAN-c for CelebA to achieve conditional image-to-image translation. The basic assumption on this translation transferring is that the two datasets share the same semantic representations (i.e., face images), the $\alpha(\cdot)$ trained on Facescrub provides a viable way to estimate the domain-specific features for the faces in the CelebA . The results of multiple identity translation on CelebA are shown in Figure~\ref{fig:celeba_intra}. We can observe that our model can still successfully translate the identity of input to the identity of conditional input with invariant domain-independent features (i.e., face emotion, face orientation).

To verify the disentanglement of domain-independent features and domain-specific features, we show the results of domain-specific feature interpolations in Figure~\ref{fig:celeba_intra_interp}.  We sample two conditional inputs $x_{B1}$ and $x_{B2}$ and obtain their domain-specific features  $x_{B1}^s$ and  $x_{B2}^s$. Then we linearly interpolate between  $x_{B1}^s$ and $x_{B2}^s$ and combine intermediary feature with domain-independent features of input $x_A$. We observe that our model can produce smooth translations through variation of  domain-specific feature while remaining the domain-independent features invariant. This indicates that DosGAN-c learns not only individual translation with specific real images, but also a generalized domain-specific feature distribution.

\begin{figure}[htb!]
	\centering
	\centerline{\includegraphics[width=8.5cm]{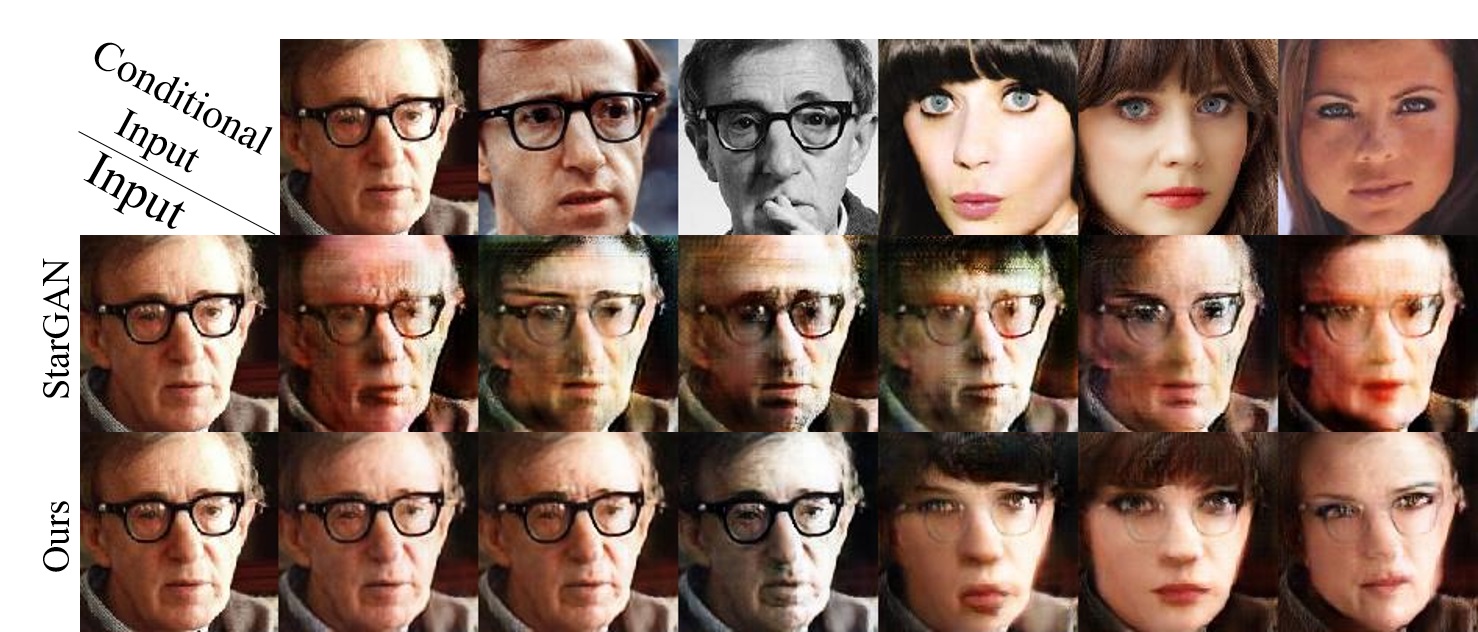}}
	\caption{Our results on multiple identity translation on $3$ identities in Facescrub with domain supervision from $528$ identities in Facescrub.}
	\label{fig:facescrub_intra_sp}
\end{figure}
\begin{table}[htb!]
\footnotesize
\setlength{\abovecaptionskip}{0.cm}
\setlength{\belowcaptionskip}{-0.cm}
\centering
\caption{Top-1 and top-5 face recognition accuracy [\%] of $3$ identities identity translation results on Facescrub.}
\label{table:face_reg_acc_sp}
\scalebox{1.0}{
\begin{tabular}{c c c c}
\toprule
    &   StarGAN \cite{choi2017stargan}    &DosGAN-c \\
\midrule
Top-1 & 6.75 &58.46   \\
Top-5 & 20.77 & 73.55 \\
\bottomrule
\end{tabular}}
\end{table}
\subsection{Adapting translation to unseen domains}\label{sec:adapting_unseen}
To further evaluate the transferability of DosGAN-c, we choose $528$ of $531$ identities in Facescrub as the dataset with domain information, and the rest $3$ identities are shuffled and used as dataset without domain information. We train the StarGAN and our DosGAN-c on $528$ identities. Then we directly test the two models on the rest $3$ identities, without any further tuning like that in Section~\ref{sec:trans_adapt¡ª_tune}. The domain codes of StarGAN for unknown $3$ identities are provided based on the prediction of classifier trained on $528$ identities. The translation examples on $3$ identities are shown in Figure~\ref{fig:facescrub_intra_sp}. The face recognition accuracy of translation results on $3$ identities are shown in Table~\ref{table:face_reg_acc_sp}. We can observe that performance of StarGAN decreases rapidly and StarGAN can not generalize its model to image translation that contains unseen domains. On the contrary, given pre-trained classifier, our model does not need any other additional information to deal with unseen identities, and shows consistent performance on image translation across different datasets.
\begin{figure}[tb!]
	\centering
	\centerline{\includegraphics[width=9cm]{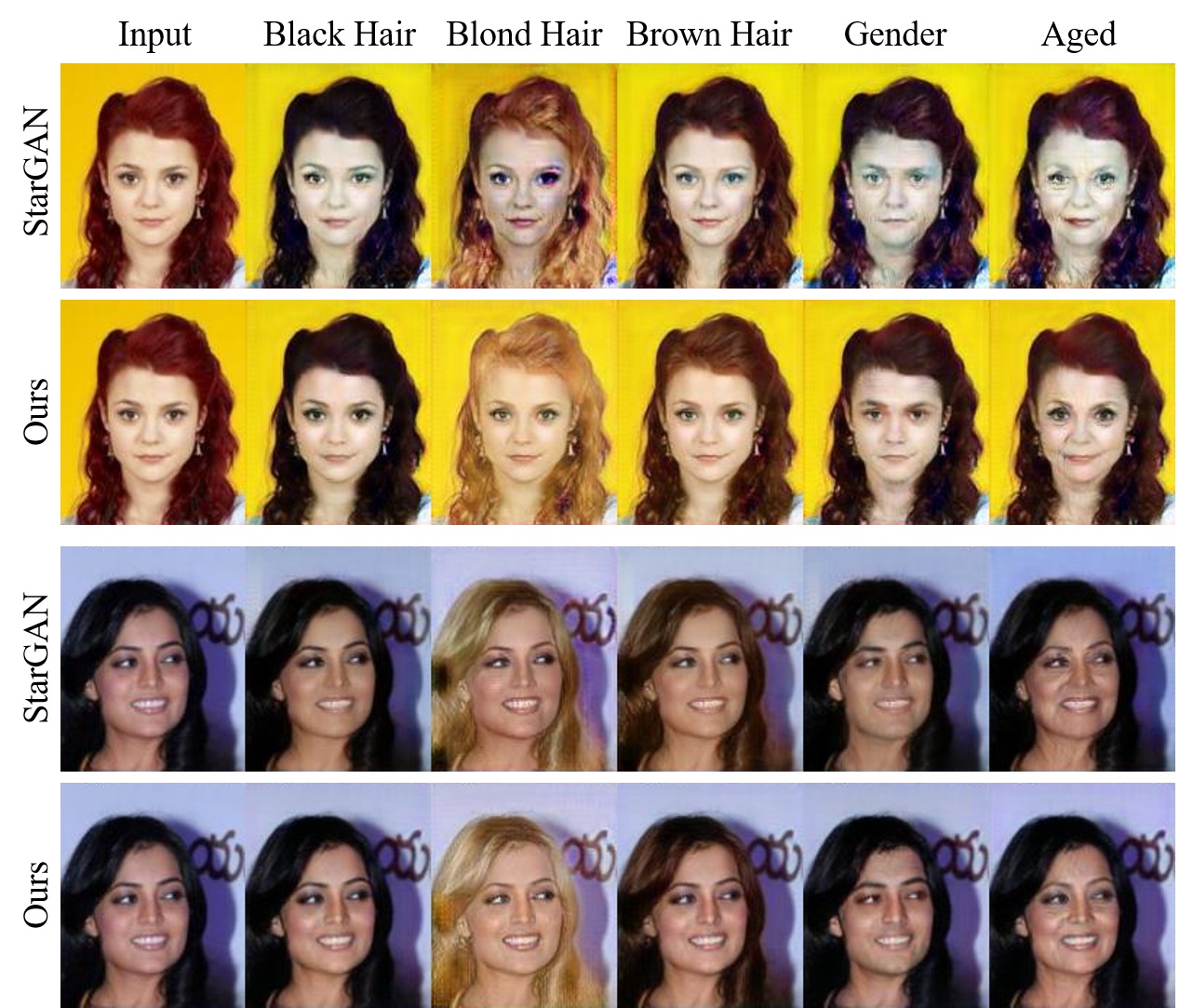}}
	\caption{Comparison between StarGAN and our DosGAN on multiple facial attribute translation. }
	\label{fig:multi_domain_trans}
\end{figure}
\begin{table}[htb!]
\footnotesize
\setlength{\abovecaptionskip}{0.cm}
\setlength{\belowcaptionskip}{-0.cm}
\centering
\caption{Classification error rate on multiple facial attribute translation.}
\label{table:clse_multidomain}
\begin{tabular}{c c c c}
\toprule
&Hair Color &Gender &Age \\
\midrule
StarGAN \cite{choi2017stargan} &19.01\%  & 11.60\%& 25.52\%\\
Ours   & 17.56\% & 11.12\%&23.79 \% \\
\bottomrule
\end{tabular}
\end{table}
\subsection{Multiple facial attribute translation}\label{sec:multi_domain-trans}
To further verify the effectiveness of our DosGAN on non-conditional setting, we compare with StarGAN on multiple facial attribute translation as shown in Figure \ref{fig:multi_domain_trans}. We can observe that our model can ensure more consistency of domain-independent features in the translated results than StarGAN. For example, in the top part of Figure \ref{fig:multi_domain_trans}, faces generated by StarGAN are found to contain more noise than our model. In addition, our model can also produce more accurate domain-specific features in the translated results than StarGAN. For example, in the bottom part of Figure \ref{fig:multi_domain_trans}, the blond hair generated by StarGAN is not very pure and has little brown color. We also show the classification error rate of translation results in Table \ref{table:clse_multidomain}. We can see that our model achieves lower classification error rate than StarGAN on three different domains, which corresponds with qualitative results.

\begin{figure*}[htb!]
	\centering
    \centerline{\includegraphics[width=17.5cm]{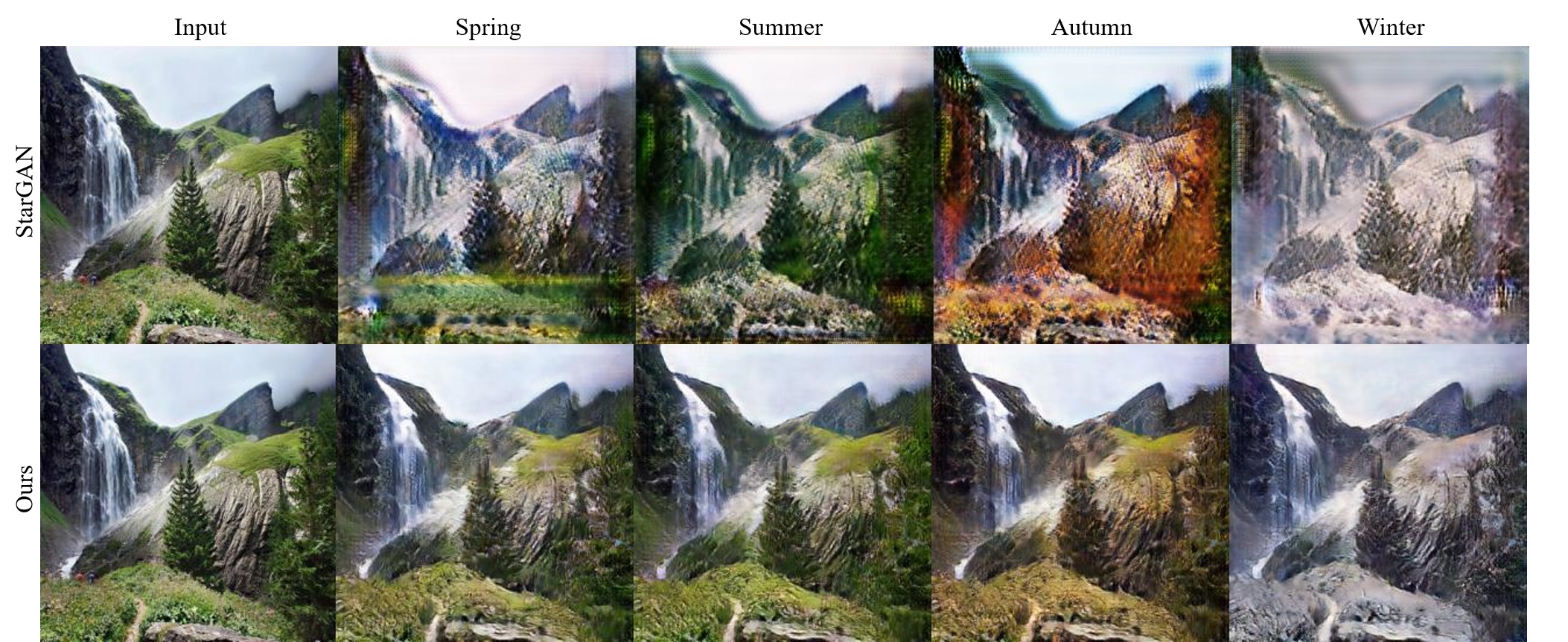}}
	\caption{Comparison between StarGAN and our DosGAN on multiple season translation.}
	\label{fig:season}
\end{figure*}

\begin{figure*}[htb!]
	\centering
    \centerline{\includegraphics[width=18.5cm]{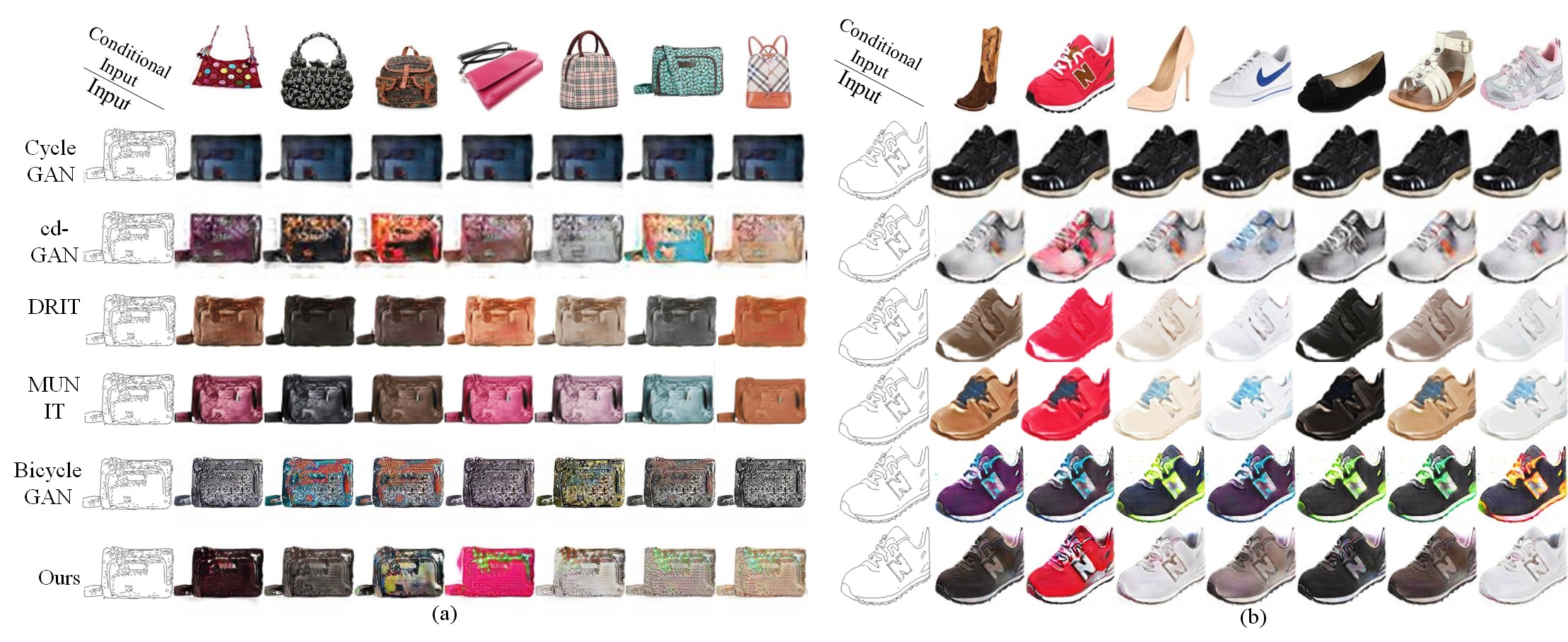}}
	\caption{Comparison among CycleGAN, cd-GAN, DRIT, MUNIT, BicycleGAN and our DosGAN-c on conditional edges$\rightarrow $handbags (a) and edges$\rightarrow $shoes (b) translation. The conditional input provides domain-specific features to the translation result.}
	\label{fig:edges-to-shoes}
\end{figure*}
\subsection{Multiple season translation}\label{sec:multi_season_trans}
We also carry out multi-domain translation on multiple season translation task. This task aims to demonstrate the effectiveness of DosGAN on generating higher resolution images and natural images. The translation results compared with StarGAN are shown in Figure \ref{fig:season}. Compared with StarGAN, we can observe that our model can translate the input images to the target domain with more accurate details and higher quality. For example, the translated results of our DosGAN have more clear edges,  precisely contain  most season features, remain other season-invariant features, such as the waterfall.  We also present the Fr¨¦chet Inception Distance (FID) \cite{NIPS2017_7240} scores of StarGAN and DosGAN in Table \ref{table:season_fid}. The FID scores verify our model's significant improvement on StarGAN.

\begin{table}[tb!]
\footnotesize
\setlength{\abovecaptionskip}{0.cm}
\setlength{\belowcaptionskip}{-0.cm}
\centering
\caption{FID scores of multiple season translation.}
\label{table:season_fid}
\scalebox{1.0}{
\begin{tabular}{c c c c}
\toprule
    &   StarGAN \cite{choi2017stargan}  & DosGAN  &Improvement\\
\midrule
FID & 81.81 & 51.14 &30.67 \\
\bottomrule
\end{tabular}}
\end{table}

\subsection{Conditional cross-domain translation results}\label{sec:ccross_domain-trans}

\begin{table*}[!tb]
\centering
\caption{PSNR [dB], SSIM, FID between ground truth images and translation images generated from paired edge images and ground truth images. The UR is the user study statistics. $\uparrow$ means the larger score is better. $\downarrow$ means the lower score is better.}
\label{table:psnr_conditional}
\begin{tabular}{ccccccccc}
\toprule
           & \multicolumn{4}{c}{edges$\rightarrow$handbags} & \multicolumn{4}{c}{edges$\rightarrow$shoes} \\
           \midrule
           & PSNR$\uparrow$           & SSIM$\uparrow$           & FID$\downarrow$   &US$\uparrow$       & PSNR$\uparrow$          & SSIM$\uparrow$          & FID$\downarrow$   &US$\uparrow$      \\
           \midrule
CycleGAN \cite{zhu2017unpaired}   &  8.45              &  0.3446      &    86.40  & 21.3\%       & 13.23              &   0.5028         & 75.54   &  23.4\%       \\
cd-GAN \cite{lin2018conditional}     & 16.05               &   0.6590         &  84.50  & 43.5\%          & 17.85              & 0.7149   &  80.73 &45.2\%      \\
DRIT \cite{lee2018diverse}       &   11.49             &  0.5139            &98.27    & 38.1\%         &  16.49             & 0.7226        &84.64  &42.7\%       \\
MUNIT \cite{huang2018multimodal}      &   14.54             &  0.6440              & 79.32 & 44.2\%           & 17.95      &  0.7353    &\textbf{70.72}    &48.7\%       \\
DosGAN-c   & \textbf{16.47}               &   \textbf{0.6620}            &  \textbf{78.67}   & \textbf{50.0\%}        &  \textbf{18.02}             &    \textbf{0.7376}           &    71.22    &  \textbf{50.0\%}   \\
\midrule
BicycleGAN \cite{zhu2017toward} (Supervised) & 16.65               &    0.6779         &  75.59   & 54.8\%        & 18.95               &  0.7605 &  70.81   & 56.9\%     \\
\bottomrule
\end{tabular}
\end{table*}
To further investigate the effectiveness of our DosGAN-c on conditional image-to-image setting, we compare our model with CycleGAN \cite{zhu2017unpaired}, DRIT \cite{lee2018diverse}, MUNIT \cite{huang2018multimodal}, BicycleGAN \cite{zhu2017toward} and cd-GAN \cite{lin2018conditional} on conditional edges$\rightarrow $handbags and edges$\rightarrow $shoes. The conditional translation results are shown in Figure \ref{fig:edges-to-shoes}. In order to measure the accuracy of domain-specific features in translation results, we present the quantitative comparison results in Table \ref{table:psnr_conditional}, where we calculate Peak Signal to Noise Ratio (PSNR), Structural Similarity (SSIM), FID score between ground truth images and translated images generated from paired edge images (providing domain-independent features) and ground truth shoe/handbag images (providing domain-specific features). From Figure \ref{fig:edges-to-shoes}, we show that our model can visually transfer the domain-specific features from the conditional input to the translation result. The quantitative results in Table \ref{table:psnr_conditional} further verify that our model's performance is better than cd-GAN, DRIT, MUNIT and comparable with supervised model BicycleGAN.

In addition to objective measurements, we carry out human perceptual study to further evaluate our results. The reviewers are given an input image and two translated images from two different methods, and they have to choose the better one without knowing which method the images are generated from. We randomly select $100$ groups of images for each compared method. The User Study (UR) statistics are also shown in Table~\ref{table:psnr_conditional}. We can observe that our model achieves the best user study result among the unsupervised image-to-image models.

\subsection{Ablation study}
\begin{table}[!tb]
\footnotesize
\setlength{\abovecaptionskip}{0.cm}
\setlength{\belowcaptionskip}{-0.cm}
\centering
\caption{Ablation study of DosGAN on multiple identity translation.}
\label{table:ablation_study}
\scalebox{1.0}{
\begin{tabular}{c c c c c}
\toprule
    &Top-1 (\%)  & Top-5 (\%)     \\
\midrule
DosGAN w/o $\ell_{\text{im}}$ &55.28 & 70.19      \\
DosGAN w/o $\ell_{\alpha,f}$ &11.16 & 13.28\\
$\text{DosGAN}_f$  &15.99 & 20.52    \\
$\text{DosGAN}_p$  &61.21 & 77.32    \\
DosGAN &94.01 & 97.98\\
\bottomrule
\end{tabular}}
\end{table}
To  verify the effectiveness of the image reconstruction loss $\ell_{\text{im}}$, domain-specific reconstruction loss $\ell_{\alpha,f}$ and pre-training strategy, we quantitatively compare DosGAN with four variants on the multiple identity translation. Two variants of DosGAN ablate $\ell_{\text{im}}$ and $\ell_{\alpha,f}$. The other one variant (briefly denoted as $\text{DosGAN}_f$) replaces $d_{feat}$ with $\alpha$ in Eqn. (\ref{eq:ds_reconstruct_loss}), omits the $\ell_{\alpha,r}$ loss and only optimizes the new $\ell_{\alpha,f}$ loss
\begin{equation}
\begin{aligned}
\ell_{\alpha,f} =\frac{1}{\vert D_A\vert}\sum_{x_{A}\in D_A}[\Vert \alpha(x_{AB}) - S_B \Vert_1].
\end{aligned}
\end{equation}

The another one variant (briefly denoted as $\text{DosGAN}_p$) optimize all network components together instead of pre-training $\alpha(\cdot)$. Without touching $\ell^{\text{total}}_{d}$, $\alpha(\cdot)$, $\beta(\cdot)$ and $g$ are optimized by a new loss $\ell^{\text{total}}_{\text{net}}$:

\begin{equation}
\begin{aligned}
\ell^{\text{total}}_{\text{net}}= \ell_{\text{GAN}}+ \ell_{\text{cls}}+\lambda_{f}\ell_{\alpha,f}+\lambda_{\text{im}}\ell_{\text{im}},
\label{eq:total_loss_generator_classifier}
\end{aligned}
\end{equation}

\begin{equation}
\begin{aligned}
\ell_{\text{cls}}  = -P(l=A|x_A;\alpha(\cdot)).
\end{aligned}
\end{equation}
As shown in Table \ref{table:ablation_study}, removing $\ell_{\text{im}}$ and  $\ell_{\alpha,f}$ from DosGAN leads to significantly worse classification accuracy rate. In addition, the $\text{DosGAN}_f$ also fails to translate images as DosGAN, which indicates that the necessity of the $\ell_{\alpha,f}$ and $\ell_{\alpha,r}$ to DosGAN's training. Without pre-training, $\text{DosGAN}_p$ also shows performance degradation compared with original model, which verifies that an explicitly formulated domain classifier can lead to better factor disentanglement of domain-specific features and domain-independent features.
\section{Conclusions}\label{sec:conclusion}
In this paper, we have presented a novel framework called  DosGAN/DosGAN-c to utilize domain information as explicit supervision for unconditional or conditional image-to-image translation. The effectiveness of our approach has been demonstrated on cross-domain and multi-domain image translation tasks. Moreover, the pre-trained domain-specific feature extractor can be transferred to other datasets without domain supervision information.

There are several interesting future directions. First, we can apply our model to more image translation tasks. Second, we would like to explore how to use our techniques for zero-shot image-to-image translation. Furthermore, we believe that our framework can be generalized to other tasks such as language and speech.

{
\bibliographystyle{unsrt}
\bibliography{Bibliography-File}
}

\end{document}